\documentclass[runningheads]{llncs}
\usepackage[T1]{fontenc}
\usepackage{graphicx,verbatim}
\usepackage{booktabs} 
\usepackage{amsmath,amssymb}
\usepackage{multirow}
\usepackage{xcolor}
\usepackage{hyperref}
\usepackage{cleveref}
\hypersetup{hidelinks, pdfauthor={}, pdftitle={}, pdfkeywords={}}
\begin{document}
\title{LoV3D: Grounding Cognitive Prognosis Reasoning in Longitudinal 3D Brain MRI via Regional Volume Assessments}
\titlerunning{LoV3D}
\begin{comment}  %% Removed for anonymized MICCAI submission
\author{First Author\inst{1}\orcidID{0000-1111-2222-3333} \and
Second Author\inst{2,3}\orcidID{1111-2222-3333-4444} \and
Third Author\inst{3}\orcidID{2222--3333-4444-5555}}
%
\authorrunning{F. Author et al.}
\institute{Princeton University, Princeton NJ 08544, USA \and
Springer Heidelberg, Tiergartenstr. 17, 69121 Heidelberg, Germany
\email{lncs@springer.com}\\
\url{http://www.springer.com/gp/computer-science/lncs} \and
ABC Institute, Rupert-Karls-University Heidelberg, Heidelberg, Germany\\
\email{\{abc,lncs\}@uni-heidelberg.de}}

\end{comment}

\author{Zhaoyang Jiang\inst{1} \and
        Zhizhong Fu\inst{2} \and
        David McAllister\inst{1} \and
        Yunsoo Kim\inst{3} \and
        Honghan Wu\inst{1}}

\authorrunning{Z. Jiang et al.}

\institute{School of Health \& Wellbeing, University of Glasgow, Glasgow, UK \\
    \email{\{3167645J, David.McAllister, Honghan.Wu\}@glasgow.ac.uk}
    \and
    School of Life Science and Technology, University of Electronic Science and Technology of China, Chengdu, China \\
    \email{zhizhong.fu@std.uestc.edu.cn}
    \and
    Institute of Health Informatics, University College London, London, UK \\
    \email{yunsoo.kim.23@ucl.ac.uk}}
\maketitle              % typeset the header of the contribution
\begin{abstract}
Longitudinal brain MRI is essential for characterizing the progression of neurological diseases such as Alzheimer's disease assessment. However, current deep-learning tools fragment this process: classifiers reduce a scan to a label, volumetric pipelines produce uninterpreted measurements, and vision-language models (VLMs) may generate fluent but potentially hallucinated conclusions. We present LoV3D, a pipeline for training 3D vision-language models, which reads longitudinal T1-weighted brain MRI, produces a region-level anatomical assessment, conducts longitudinal comparison with the prior scan, and finally outputs a three-class diagnosis (Cognitively Normal, Mild Cognitive Impairment, or Dementia) along with a synthesized diagnostic summary. The stepped pipeline grounds the final diagnosis by enforcing label consistency, longitudinal coherence, and biological plausibility, thereby reducing the risks of hallucinations. The training process introduces a clinically-weighted Verifier that scores candidate outputs automatically against normative references derived from standardized volume metrics, driving Direct Preference Optimization without a single human annotation. On a subject-level held-out ADNI test set (479 scans, 258 subjects), LoV3D achieves 93.7\% three-class diagnostic accuracy with zero non-adjacent errors, 97.2\% two-class accuracy (+4\% over SOTA), and 82.6\% region-level anatomical classification accuracy (+33.1\% over VLM baselines). Removing anatomical grounding from the encoder drops accuracy to 92.5\% and introduces a critical CN$\leftrightarrow$Dementia error. Zero-shot transfer yields 95.4\% on MIRIAD (100\% Dementia recall) and 82.9\% three-class accuracy on AIBL, confirming high generalizability across sites, scanners, and populations. Code is available at \url{https://github.com/Anonymous-TEVC/LoV-3D}.

\keywords{3D vision-language model \and longitudinal brain MRI \and structured reasoning \and Alzheimer's disease \and preference optimization}
% Authors must provide keywords and are not allowed to remove this Keyword section.

\end{abstract}

\section{Introduction}
\label{sec:intro}

Alzheimer's disease (AD) is the leading cause of dementia, and longitudinal brain MRI is central to tracking its progression~\cite{jack2018nia}. Serial T1-weighted scans reveal characteristic atrophy in the hippocampus, entorhinal cortex, and temporal neocortex, a trajectory that informs diagnosis, staging, and treatment planning. However, when a neuroradiologist reads a follow-up scan, the resulting report is far more than a diagnostic label. It is a layered document: anatomical observations grounded in the image, clinical context from cognitive testing, comparison with the prior scan, and a synthesized impression. The reasoning process, not just its conclusion, is what gives the report clinical value. Therefore, automating this assessment demands more than classification; it requires structured reasoning that can be traced, questioned, and verified.

Current tools are limited because they collapse this process into a single output. Deep learning classifiers~\cite{wen2020convolutional,chen2019med3d} reduce a scan to a diagnostic label, discarding anatomical specificity. Volumetric pipelines such as FreeSurfer~\cite{fischl2012freesurfer} deliver precise measurements but not reasoning. Vision-language models~\cite{li2023llava,tu2024towards} generate fluent narratives, yet can describe hippocampal atrophy in a patient whose hippocampus is normal, and no algorithm can detect such errors from free text alone. In every case, the diagnostic process is either absent or unverifiable.

Recent work has advanced each component independently. Prediction models now incorporate longitudinal 3D data~\cite{chen2024longformer}, multimodal clinical records~\cite{lin2025vision}, and video-based VLM architectures~\cite{dang2026forecasting}, yet their outputs remain labels with no diagnostic reasoning attached. Report generation for brain MRI has begun to emerge~\cite{chiumento2024leveraging,lee2024improving}, but current approaches either decouple vision from language (feeding FreeSurfer measurements to an LLM that never sees the image) or remain limited to 2D slices. Generalist 3D VLMs~\cite{wu2025towards,bai2024m3d} enable language generation from volumetric data but lack longitudinal reasoning and structured outputs. To our knowledge, no existing system jointly reads longitudinal 3D brain MRI end-to-end, produces structured diagnostic reasoning, and trains that reasoning through automated verification.

We observe that this gap closes once the output is designed for verifiability. If the model produces not free text but structured JSON whose fields are logically linked, hallucinations become detectable by code. The key insight is that the same structure that enables detection also enables training: a clinically-weighted Verifier can score candidate outputs automatically, constructing preference pairs for direct preference optimization (DPO) without a single human annotation. This closed loop, from output design through automated verification to preference alignment, is the core principle of LoV3D. Our contributions are: (1)~a structured verifiable output format where reasoning-label consistency, longitudinal coherence, and biological plausibility constraints are checkable by code; (2)~a normative Z-score model with soft tolerance zones and a clinically-weighted Verifier that drives DPO without human labels; and (3)~comprehensive evaluation on ADNI with zero-shot cross-site transfer to MIRIAD and AIBL, demonstrating robust generalization across sites, scanners, and populations.

%==============================================================================
\section{Method}
\label{sec:method}

\Cref{fig:architecture} gives an end-to-end overview. We now describe the encoder-projector-LLM architecture (\Cref{sec:architecture}) and the verifiable output design that enables automated training (\Cref{sec:output}).

\subsection{Architecture Overview}
\label{sec:architecture}
\begin{figure}[t]
\centering
\includegraphics[width=\textwidth]{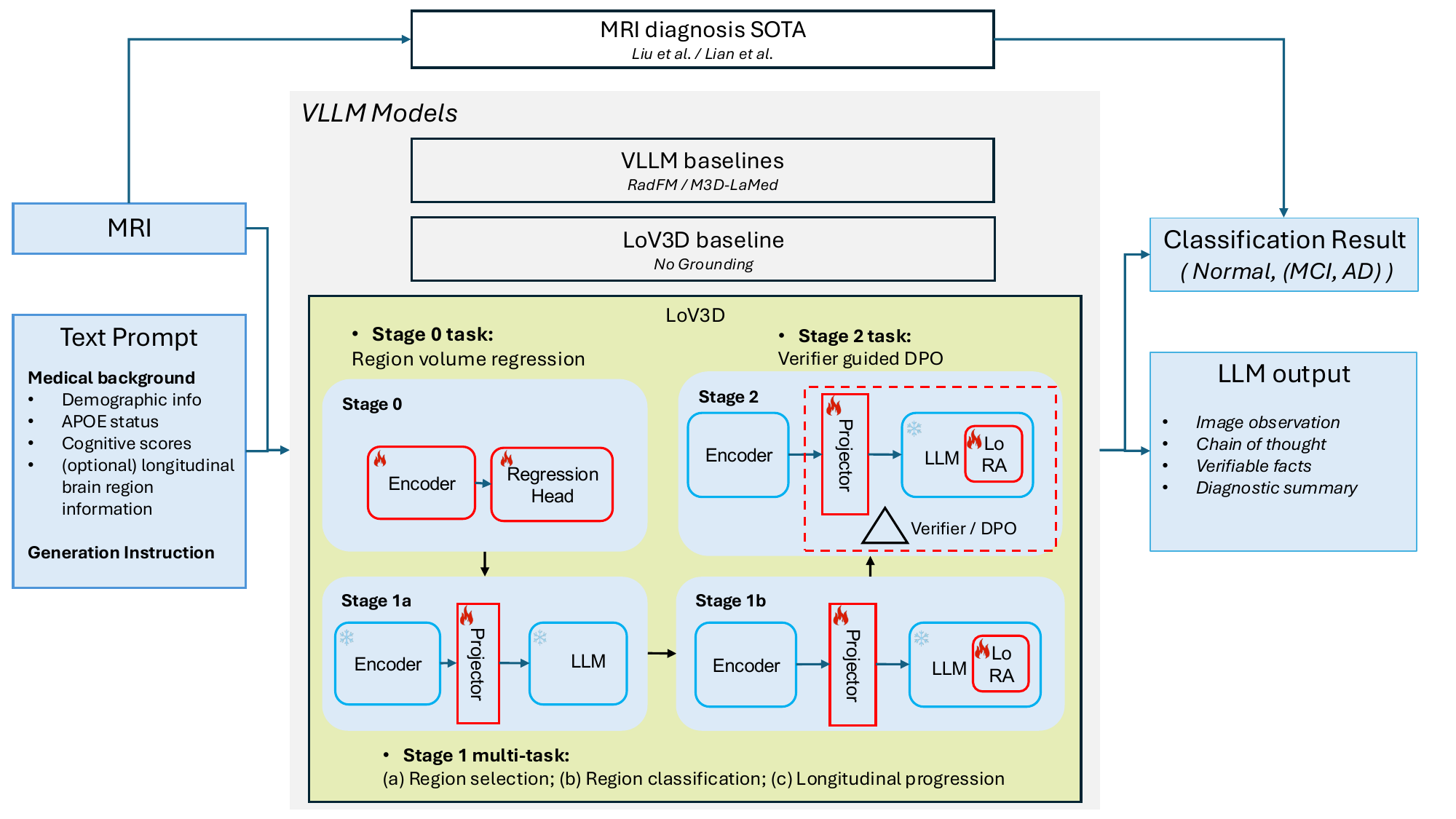}
\caption{Model paradigms and the LoV3D training pipeline.
\textbf{Top}: Published MRI classifiers produce only a two-class diagnostic label from MRI inputs. \textbf{Middle and Bottom}: multimodal VLM models that take a MRI and text prompt, and produce an output including (3-class) classification and a detailed LLM output. Specifically,
\textbf{Middle}: 3D medical VLMs (RadFM, M3D-LaMed) and an ungrounded ablation (same pipeline, encoder warmed via diagnosis classification instead of regional regression). \textbf{Bottom}: LoV3D training %receives both MRI and a text prompt carrying medical background and, for follow-ups, prior regional information.
introduces a three-stage learning paradigm to ground the VLM's cognitive prognosis reasoning. Stage~0 warms the encoder via a regional volume regression task; Stages~1a--b learn structured clinical reasoning through projector alignment and LoRA SFT for a multi-task learning (selecting relevant brain regions; classifying region deterioration categories; the longitudinal progression classification); Stage~2 refines outputs via Verifier-guided DPO (dashed border marks optimized modules) for a more reasonable diagnostic summary generation.
LoV3D produces both a classification result and structured LLM output (right) comprising imaging observations, chain-of-thought reasoning, verifiable facts, and a diagnostic summary. 
Fire/snowflake icons denote trainable/frozen modules.}
\label{fig:architecture}
\end{figure}

LoV3D connects a 3D visual encoder to a large language model through a learnable projector (Stages~1a--2 in \Cref{fig:architecture}). We build this pipeline from modular components rather than fine-tuning existing 3D medical VLMs, because current options cannot produce the structured outputs our task requires: both RadFM and M3D-LaMed achieve 0\% valid JSON in zero-shot evaluation (\Cref{sec:main_results}), indicating that general-purpose 3D medical pretraining does not confer the instruction-following capacity needed for structured clinical reporting. A MONAI ResNet-50~\cite{chen2019med3d} truncated after \texttt{layer3} produces a feature map in $\mathbb{R}^{1024 \times 16 \times 16 \times 16}$, pooled to $8{\times}8{\times}8$ and reshaped into 512 visual tokens. We choose a CNN over vision transformers because the encoder is warmed up on a limited number of baseline scans, a regime where data-hungry architectures such as ViT and Swin readily overfit. A two-layer MLP with GELU activation projects each token from 1024 to $d{=}5120$ dimensions, matching the embedding space of Qwen-2.5-14B~\cite{yang2024qwen2}. These visual tokens are inserted into the tokenized text prompt and processed jointly by the LLM with LoRA adapters~\cite{hu2022lora}.

\begin{sloppypar}
The text prompt carries demographics, APOE~$\varepsilon$4 status, cognitive scores (MMSE, CDR-SB), and for follow-up visits, prior anatomical labels from FreeSurfer analysis of the previous scan along with longitudinal history. Crucially, the current scan's FreeSurfer measurements serve exclusively as Verifier ground truth and are never shown to the model; the model must derive its anatomical assessment from the 3D image itself.

\end{sloppypar}

\subsection{Verifiable Output Design: multi-task learning}
\label{sec:output}

The key to closing the training loop is an output format whose correctness can be checked by code.
LoV3D produces a JSON object whose fields fall into two categories (\Cref{fig:architecture}, right): qualitative fields that carry free-text clinical reasoning, and verifiable fields constrained by explicit logical relationships. Following the JSON, the model generates a [Diagnostic Summary] paragraph that synthesizes findings into natural language.

The JSON follows a reasoning-first ordering: the model first generates imaging observations and clinical integration text, articulating what it sees and how clinical scores relate to the visual findings, before committing to categorical predictions for anatomical assessment and diagnosis. This mirrors clinical practice, requiring deliberation before diagnosis, and during autoregressive generation it means the model has already produced its evidence by the time it must declare a verdict.

The verifiable fields encode three checkable constraints. \textbf{C1~region selection}: each region labeled as abnormal must be referenced in the model's reasoning text, and progression must be cited whenever a severity threshold is crossed. 
\textbf{C2~region classification}: since neurodegeneration is irreversible, any current label that is two or more severity levels milder than the prior label is flagged as implausible. \textbf{C3~Longitudinal progression classification}: the predicted change direction (stable, progressive atrophy, or progressive enlargement) and the threshold-crossing flag must be mutually consistent for each region. Together, these constraints enable model evaluation as an algorithmic procedure, thereby making the entire DPO loop possible.

\subsection{Normative Z-Score Model}
\label{sec:normative}

The verifiable output requires categorical anatomical labels as ground truth. We derive these from FreeSurfer volumetric measurements through a normative Z-score model fitted on cognitively normal training subjects. For each AD-signature region $r \in \mathcal{R}$, the Z-score
\begin{equation}
z_r = \frac{v_r - (\alpha_r + \beta_r^{\text{age}} \cdot \text{age} + \beta_r^{\text{sex}} \cdot \mathbb{1}[\text{male}])}{\sigma_r}
\label{eq:zscore}
\end{equation}
quantifies deviation from age-sex-adjusted norms, where $v_r$ is the ICV-normalized volume and $\sigma_r$ the residual standard deviation. Fitting exclusively on training-split CN subjects prevents leakage. Z-scores are discretized into three severity levels: normal ($z_r > -0.5$), mild atrophy ($-1.5 < z_r \leq -0.5$), and severe atrophy ($z_r \leq -1.5$), with mirrored thresholds for ventricular enlargement. The $-1.5$ SD boundary follows the established impairment cutoff from MCI diagnosis~\cite{petersen1999mild}; the $-0.5$ SD boundary captures early volume loss, consistent with the 20--40\% prevalence of preclinical amyloid pathology in cognitively normal elderly~\cite{jack2010hypothetical}. This three-level scheme offers finer granularity than the binary normal/abnormal classification used in most prior work while maintaining reliable inter-class separability; we verify that the model is robust to threshold choice by re-evaluating under five configurations (mild cutoff $-0.3$ to $-0.7$, severe $-1.0$ to $-2.0$): region accuracy varies by less than 3.3~pp across the three central settings, and diagnostic accuracy is entirely unaffected since diagnosis labels come from clinical assessment rather than Z-scores.

A subtlety arises near boundaries. A Z-score of $-0.52$ and one of $-0.48$ may land on different sides of a threshold despite negligible volumetric difference, and FreeSurfer's own segmentation variability, particularly in atrophied tissue where grey--white boundaries become ambiguous, further compounds this uncertainty. If the Verifier treats boundary cases as hard errors, the resulting DPO signal becomes noisy. We introduce soft tolerance zones of $\pm 0.25$ Z around each boundary: within these zones, the Verifier accepts adjacent labels with partial credit scaled by the model's expressed confidence.

\subsection{Clinically-Weighted Verifier}
\label{sec:verifier}

With structured outputs and normative labels in hand, the Verifier scores each candidate against FreeSurfer-derived ground truth that the model never sees:
\begin{equation}
S_{\text{verifier}} = M(\hat{d}, d^*) \cdot \sum_{c \in \mathcal{C}} \lambda_c \, S_c
\label{eq:verifier}
\end{equation}
where $\mathcal{C} = \{\text{anat, dx, long, reason, summary}\}$, $M$ is a global clinical multiplier (${\times}2.0$ for non-adjacent diagnostic errors, ${\times}1.5$ for adjacent) that discounts the entire output because a catastrophic misdiagnosis signals systemic reasoning failure beyond the diagnosis field alone, and $\boldsymbol{\lambda} = (0.25, 0.25, 0.20, 0.15, 0.15)$ for follow-up visits (anatomy and diagnosis each receive 0.35 for baselines without longitudinal data).

$S_{\text{anat}}$ compares predicted labels against normative ground truth with tolerance-zone scoring (\Cref{sec:normative}), weighting hippocampus ($w{=}1.2$) and entorhinal cortex ($w{=}1.1$) for their diagnostic primacy in AD:
\begin{equation}
S_{\text{anat}} = \frac{\textstyle\sum_{r \in \mathcal{R}} w_r \cdot s_r}{\textstyle\sum_{r \in \mathcal{R}} w_r}
\label{eq:anat}
\end{equation}
$S_{\text{dx}}$ penalizes non-adjacent errors (CN$\leftrightarrow$Dementia) twice as heavily as adjacent ones. $S_{\text{long}}$ rewards correct change direction and threshold-crossing detection, with asymmetric penalties for missed impossible reversals. $S_{\text{reason}}$ checks that abnormal regions appear in reasoning text, progression is cited for threshold crossings, and fields are internally coherent. $S_{\text{summary}}$ evaluates factual alignment between the diagnostic summary and the structured JSON.

\subsection{Verifier-Guided Preference Optimization}
\label{sec:dpo}

The Verifier's automated scoring enables preference optimization without human annotation. For each training sample, $K{=}4$ candidate responses are generated with stochastic sampling (temperature 0.7) and scored. The best- and worst-scoring candidates form the chosen-rejected pair for DPO~\cite{rafailov2023direct}:
\begin{equation}
\mathcal{L}_{\text{DPO}} = -\mathbb{E}\!\left[\log \sigma\!\left(\beta \!\left[\log \frac{\pi_\theta(y_w | x)}{\pi_{\text{ref}}(y_w | x)} - \log \frac{\pi_\theta(y_l | x)}{\pi_{\text{ref}}(y_l | x)}\right]\right)\right]
\label{eq:dpo}
\end{equation}
with $\beta {=} 0.1$. When the best candidate's score falls below a quality threshold, the ground-truth response substitutes as $y_w$, providing a learning signal for the model's weakest cases.

This is where the structured output design pays its largest dividend. A free-text Verifier would face a fundamental ambiguity: NLG metrics like ROUGE reward lexical overlap, not clinical correctness. ``Mild hippocampal atrophy'' and ``no hippocampal atrophy'' share most of their tokens but carry opposite clinical meaning. LLM-as-judge alternatives are non-deterministic and expensive, reintroducing a bottleneck that defeats the purpose of automated scoring. Structured verification sidesteps both issues: label comparison is deterministic, inter-field constraint checking is cheap, and the resulting scores directly reflect clinical accuracy rather than surface-level text similarity. We verify this robustness by regenerating all preference pairs under four alternative weight configurations (equal components, anatomy-heavy, diagnosis-heavy, equal regions): 91--97\% of chosen-rejected pairs remain identical and Kendall's $\tau$ never falls below 0.94, confirming that the discrete scoring events (correct vs.\ incorrect label) dominate over continuous weight differences.

\subsection{Training Pipeline}
\label{sec:pipeline}

Training proceeds through four stages (\Cref{fig:architecture}). \textbf{Stage~0} warms the 3D encoder via multi-task volume regression on baseline-only scans, after which the encoder is frozen and transferred. \textbf{Stage~1a} aligns the projector with the frozen LLM using causal LM loss, teaching the projector to map 3D visual tokens into the LLM's embedding space. \textbf{Stage~1b} jointly trains the projector and LoRA adapters~\cite{hu2022lora} with differential learning rates, tuning the LLM to produce well-structured clinical outputs. \textbf{Stage~2} merges the Stage~1b LoRA, applies a fresh adapter, and runs Verifier-guided DPO. Each stage builds on the previous one: the encoder learns anatomy, the projector learns alignment, the LoRA learns structured reasoning, and DPO learns to prefer clinically accurate, internally consistent outputs.

%==============================================================================
\section{Experiments}
\label{sec:experiments}

\subsection{Setup}
\label{sec:setup}

We train and evaluate on ADNI~\cite{jack2008alzheimers}, which provides 8{,}114 T1-weighted scans from 2{,}575 subjects, skull-stripped, registered to MNI152, resampled to $128^3$, and intensity-normalized. After filtering for complete FreeSurfer labels, patient-level stratified splitting yields 3{,}993 / 525 / 479 train / val / test scans (2{,}059 / 258 / 258 subjects). The test set contains 163 baseline and 316 follow-up visits (38.0\% CN, 48.0\% MCI, 14.0\% Dementia).

We compare against three categories of baselines: (i)~a ResNet-50 with a linear head using the same backbone and data split; (ii)~published binary AD vs CN classifiers on ADNI~\cite{liu2018landmark,lian2020hfcn,zhang2022single}; and (iii)~two 3D medical VLMs evaluated zero-shot---RadFM~\cite{wu2025towards} and M3D-LaMed~\cite{bai2024m3d}. The primary metrics are three-way diagnostic accuracy (DX), macro F$_1$, Cohen's weighted $\kappa$, binary AD/CN accuracy, and mean region accuracy. For the stage-wise ablation we additionally report per-severity accuracy, BLEU-4 and ROUGE-L for the diagnostic summary, reasoning-label consistency (Reasoning F$_1$), longitudinal direction accuracy, and false abnormal/severe rates. The LLM backbone is Qwen-2.5-14B in bfloat16 with LoRA (rank~16, $\alpha{=}32$). All training runs on a single A100-80GB GPU.

\subsection{Baseline Comparison}
\label{sec:main_results}

\begin{table}[t]
\centering
\caption{Comparison on the ADNI test set (258 subjects). DX = three-way diagnostic accuracy, F$_1$ = macro-averaged F$_1$, $\kappa$ = Cohen's weighted kappa, Region = mean anatomical accuracy across five AD-signature regions, AD/CN = binary accuracy on the CN and Dementia subset (MCI subjects excluded, MCI predictions mapped to CN). $^*$Same MONAI ResNet-50 backbone with a linear head, trained on the same split. $^\dagger$VLM baselines produce no valid JSON; metrics are from heuristic extraction. $^\ddagger$Published results on different ADNI subsets; shown for reference. LoV3D (no-grounding) uses the full LoV3D pipeline but replaces Stage~0 regional volume regression with three-class diagnosis classification. LoV3D is evaluated on all 479 test scans (258 subjects, multiple visits).}
\label{tab:main}
\small
\begin{tabular}{@{}lccccc@{}}
\toprule
Method & DX (\%) & F$_1$ & $\kappa$ & Region (\%) & AD/CN (\%) \\
\midrule
\multicolumn{6}{@{}l}{\textit{Classification baselines}} \\
Liu et al.~\cite{liu2018landmark}$^\ddagger$ & --- & --- & --- & --- & 91.1 \\
Lian et al.~\cite{lian2020hfcn}$^\ddagger$ & --- & --- & --- & --- & 90.3 \\
Zhang et al.~\cite{zhang2022single}$^\ddagger$ & --- & --- & --- & --- & 93.2 \\
ResNet-50 (ours)$^*$ & 58.9 & 58.1 & .461 & --- & 87.8 \\

\midrule
\multicolumn{6}{@{}l}{\textit{3D medical VLM baselines}} \\
RadFM~\cite{wu2025towards}$^\dagger$ & 17.5 & --- & --- & 41.4 & --- \\
M3D-LaMed~\cite{bai2024m3d}$^\dagger$ & 38.2 & --- & --- & 49.5 & --- \\
\midrule
LoV3D (no-grounding) & 92.5 & 92.0 & .891 & 80.7 & 96.4 \\
LoV3D & \textbf{93.7} & \textbf{93.3} & \textbf{.911} & \textbf{82.6} & \textbf{97.2} \\
\bottomrule
\end{tabular}
\end{table}

\Cref{tab:main} compares LoV3D against dedicated 3D CNN classifiers and generalist 3D medical VLMs. Classification models are limited to producing a single diagnostic label without any reasoning or report, so they cannot address our full task. Generalist 3D VLMs are architecturally capable of structured reporting, yet as we show below, existing models cannot reliably perform it on longitudinal brain MRI.

Even on the diagnostic sub-task that classifiers are designed for, three-way classification (CN vs.\ MCI vs.\ Dementia) from MRI alone is inherently difficult because the structural differences between MCI and the other two stages are subtle and highly overlapping. As a result, most prior work simplifies the problem to binary AD vs.\ CN, achieving 90--93\% on ADNI subsets~\cite{liu2018landmark,lian2020hfcn,zhang2022single}. A ResNet-50 with a linear head, using the same backbone and data split as LoV3D but trained end-to-end, confirms this difficulty: it reaches only 58.9\% three-way accuracy ($\kappa{=}0.461$) and 87.8\% binary. By integrating clinical metadata (cognitive scores, APOE status) with visual features through structured LLM reasoning, LoV3D achieves 93.7\% three-way accuracy ($\kappa{=}0.911$). Its confusion matrix contains zero CN$\leftrightarrow$Dementia confusions across all 479 test scans; every error falls between adjacent categories (CN$\leftrightarrow$MCI or MCI$\leftrightarrow$Dementia). Under the standard binary protocol~\cite{lian2020hfcn}, restricting evaluation to the 249 CN and Dementia scans and mapping MCI predictions to CN, LoV3D yields 97.2\% accuracy.

While classifiers at least produce usable diagnoses, the generalist VLM baselines struggle with even this basic requirement: neither produces a single valid JSON output. RadFM echoes the prompt schema without analyzing the image; M3D-LaMed generates short free-text narratives ignoring the structured output instruction. Even after generous heuristic extraction of DX labels from their free text, RadFM reaches only 17.5\% three-way DX with near-total Dementia bias (86.8\% Dementia recall vs.\ 2.9\% MCI), and M3D-LaMed 38.2\% with the opposite bias (89.5\% CN recall, 0\% Dementia). These results confirm that general-purpose 3D medical pretraining does not confer the instruction-following capacity needed for structured clinical reporting.

To isolate the contribution of anatomical grounding, we evaluate an ablation (LoV3D, no-grounding) that replaces Stage~0 regional volume regression with three-class diagnosis classification while keeping the full VLM pipeline identical through Stages~1a--2. This variant reaches 92.5\% three-way DX ($\kappa{=}0.891$) and 80.7\% region accuracy---already far above both CNN classifiers and generalist VLM baselines---but introduces one CN$\to$Dementia confusion, the only non-adjacent error across all evaluated models. The consistent gains from grounding (+1.2~pp DX, +1.9~pp region, +0.020~$\kappa$) and the elimination of all critical errors confirm that region-level encoder pretraining instills anatomical sensitivity that propagates through the entire pipeline.

In summary, classifiers achieve reasonable diagnosis but cannot reason, and generalist VLMs can generate text but lack the task-specific capacity for reliable structured reporting. LoV3D addresses all facets: 93.7\% three-way DX, 82.6\% region accuracy, 100\% JSON validity, and diagnostic summaries (ROUGE-L{=}0.763). The no-grounding ablation further confirms that anatomical grounding is essential not merely for accuracy but for clinical safety---eliminating the sole non-adjacent diagnostic error. We next analyze how each training stage contributes to this result.

\subsection{Ablation Study: Stage-Wise Analysis}
\label{sec:ablation}

\begin{table}[t]
\centering
\caption{Stage-wise progression on the ADNI test set (479 scans). Each stage builds on the previous checkpoint. Region = mean accuracy across five AD-signature regions. F.Abn./F.Sev.\ = false abnormal/severe rate on normal anatomy ($\downarrow$ = lower is better). Reasoning completeness is 100\% and hallucination rate ${<}$1\% at all stages.}
\label{tab:ablation}
\small
\begin{tabular}{@{}l ccc c cc cc@{}}
\toprule
& \multicolumn{3}{c}{Diagnosis} & & \multicolumn{2}{c}{Report Quality} & \multicolumn{2}{c}{Clinical Safety} \\
\cmidrule(lr){2-4}\cmidrule(lr){6-7}\cmidrule(lr){8-9}
Stage & Acc & F$_1$ & $\kappa$ & Region & BLEU-4 & ROUGE-L & F.Abn$\downarrow$ & F.Sev$\downarrow$ \\
\midrule
1a (projector) & 89.1 & 88.0 & .851 & 79.7 & .431 & .635 & 7.2 & 6.3 \\
1b (+ LoRA)    & 93.3 & 92.5 & .905 & 81.6 & .354 & .558 & 7.1 & 4.1 \\
2 (+ DPO)      & \textbf{93.7} & \textbf{93.3} & \textbf{.911} & \textbf{82.6} & \textbf{.584} & \textbf{.763} & \textbf{5.0} & \textbf{2.2} \\
\bottomrule
\end{tabular}
\end{table}

\Cref{tab:ablation} traces how each training stage addresses a distinct limitation. Stage~1a aligns the projector with the frozen LLM, already achieving 89.1\% diagnostic accuracy and 79.7\% region accuracy; all 52 classification errors are adjacent (CN$\leftrightarrow$MCI or MCI$\leftrightarrow$Dementia), with zero CN$\leftrightarrow$Dementia confusions---a clinically essential property that persists through every subsequent stage. Stage~1b introduces LoRA adapters, raising DX to 93.3\% ($\kappa{=}.905$). The gains concentrate where the frozen LLM struggled most: MCI, whose anatomical boundaries with CN and Dementia are subtlest, improves by 4.7~pp in F$_1$ (88.5$\to$93.2), and follow-up accuracy jumps from 88.5\% to 94.7\%, exceeding baseline-visit accuracy (90.4\%)---confirming that LoRA enables the LLM to exploit prior anatomical labels for longitudinal reasoning. Stage~2 applies Verifier-guided DPO. The diagnostic gain is focused on the hardest category (Dementia F$_1$: 89.4$\to$91.6), but DPO's most revealing effect is on report quality: BLEU-4 rises from .354 to .584 (+65\%) and ROUGE-L from .558 to .763 (+37\%), recovering and surpassing the quality that Stage~1b's task-specific training had \emph{decreased} relative to 1a (ROUGE-L: .635$\to$.558). This non-monotonic trajectory exposes a fundamental tension: SFT sharpens classification at the cost of linguistic diversity, a trade-off that preference optimization uniquely resolves by selecting for outputs that are both clinically correct and well-formed.

DPO's clinically-weighted training signal produces targeted improvements where the stakes are highest. The false severe rate---overcalling severe pathology in normal tissue---drops from 4.1\% to 2.2\% ($-$46\%), and Dementia specificity reaches 99.0\%, confirming that the model almost never assigns the most severe diagnosis incorrectly. Per-region analysis reveals a difficulty ordering that persists across all stages: ventricles are easiest to assess (90.4\%, $\kappa{=}.842$), reflecting their large absolute volume changes, while the entorhinal cortex is the most challenging (75.2\% at Stage~1a) yet shows the largest DPO improvement (77.5\%$\to$79.5\%, $\kappa$: .721$\to$.745)---validating that the Verifier's elevated weight ($w{=}1.1$) on this diagnostically critical but segmentation-noisy region provides effective training signal where it is needed most. The per-severity breakdown identifies the principal remaining bottleneck: mild atrophy, the earliest and most clinically actionable stage, reaches 67.1\% accuracy (up from 57.9\% at 1a), with DPO contributing +2.9~pp over Stage~1b via the tolerance-zone scoring that resolves boundary cases between normal and mildly atrophied tissue. Reasoning F$_1$ reaches 87.1\%, longitudinal direction accuracy 88.2\% across 316 follow-up scans, and diagnostic ECE remains below 0.04---collectively indicating well-calibrated, internally consistent outputs across all stages.

\subsection{External Validation}
\label{sec:cross_site}

A central question is whether performance transfers across sites and scanners. We evaluate LoV3D zero-shot---without fine-tuning, domain adaptation, or normative model refitting---on two independent datasets (\Cref{tab:cross_site}).

\paragraph{MIRIAD (binary: AD vs HC).}
MIRIAD~\cite{malone2013miriad} contains 69 subjects (46~AD, 23~HC; 523 scans) from a single 1.5T scanner at UCL London, with substantially sparser metadata than ADNI (no APOE, limited MMSE/CDR). LoV3D achieves 95.4\% per-scan accuracy: all 346 Dementia scans are classified correctly (100\% recall), while the 24 errors are exclusively CN$\to$MCI confusions---an inherent artifact of mapping a three-class model onto a binary protocol. The best published results on MIRIAD as a pure external test set are 95.7\% balanced accuracy~\cite{yee2021construction} and 93.6\% accuracy~\cite{lu2022practical}, both binary classifiers without this mapping ambiguity.

\paragraph{AIBL (three-class: CN vs MCI vs Dementia).}
AIBL~\cite{ellis2009australian} presents a harder challenge: an Australian cohort (621 subjects, 989 scans) acquired on different scanners with a different demographic profile, requiring the full three-way classification. LoV3D achieves 82.9\% accuracy (balanced 74.2\%), surpassing the strongest published baselines---Lteif et al.~\cite{lteif2024disease} (73.4\%) and Batool et al.~\cite{batool2026higher} (76.4\%, balanced 61.4\%)---by over 6 percentage points despite operating zero-shot. Only 2 of 989 scans incur a critical CN$\leftrightarrow$Dementia error; follow-up scans (87.2\%) outperform baseline visits (80.4\%), confirming that longitudinal context provides additional discriminative signal. That LoV3D exceeds dedicated domain-generalization classifiers trained on NACC without any adaptation suggests the visual encoder has learned scanner-invariant anatomical representations rather than site-specific artifacts.

\begin{table}[t]
\centering
\caption{Cross-site generalization. LoV3D is evaluated zero-shot (no fine-tuning) on two independent datasets. Prior work results are reproduced from published numbers on the same datasets used as pure external test sets. MIRIAD baselines were trained on ADNI; AIBL baselines were trained on NACC.}
\label{tab:cross_site}
\small
\begin{tabular}{@{}llcccc@{}}
\toprule
Dataset & Method & Classes & Subjects & Acc (\%) & Bal.\ Acc (\%) \\
\midrule
\multirow{3}{*}{MIRIAD} & Yee et al.~\cite{yee2021construction} & 2 & 69 & --- & 95.7 \\
 & Lu et al.~\cite{lu2022practical} & 2 & 69 & 93.6 & 94.9 \\
 & LoV3D (ours) & 2 & 69 & 95.4 & 93.2 \\
\midrule
\multirow{3}{*}{AIBL} & Lteif et al.~\cite{lteif2024disease} & 3 & 661 & 73.4 & --- \\
 & Batool et al.~\cite{batool2026higher} & 3 & 661 & 76.4 & 61.4 \\
 & \textbf{LoV3D (ours)} & \textbf{3} & \textbf{621} & \textbf{82.9} & \textbf{74.2} \\
\bottomrule
\end{tabular}
\end{table}

%==============================================================================
\section{Conclusion}
\label{sec:conclusion}
LoV3D demonstrates the utility of anatomical verification based grounding in enhancing cognitive prognosis reasoning of medical VLMs: the structured output quantifying brain region deterioration makes hallucinations detectable and makes preference optimization feasible without human annotation. On ADNI the model achieves 93.7\% three-class diagnostic accuracy ($\kappa${=}0.911) with 82.6\% region accuracy and zero non-adjacent errors; Verifier-guided DPO improves report quality by 65\% (BLEU-4) and reduces false severe labels by 46\%. Zero-shot transfer yields 95.4\% on MIRIAD and 82.9\% on AIBL without any domain adaptation. Limitations include reliance on FreeSurfer-derived ground truth, restriction to T1-weighted MRI, and absence of amnestic vs.\ non-amnestic MCI distinction. The principle extends beyond neuroimaging: any domain where output correctness is critical and expensive to verify manually---from radiology to pathology to longitudinal oncology---stands to benefit from the same insight: design for verifiability first, and the training loop follows.

\bibliographystyle{splncs04}
\bibliography{Reference}

\end{document}